\documentclass[runningheads]{llncs}

\usepackage{eccv}

\usepackage{eccvabbrv}

\usepackage{graphicx}
\usepackage{booktabs}
\usepackage{graphicx}
\usepackage{amsmath}
\usepackage{booktabs}
\usepackage{algorithm}
\usepackage{algorithmic}
\usepackage{diagbox}
\usepackage{algorithmic}
\usepackage{multicol}
\usepackage{multirow}
\usepackage{booktabs}
\usepackage{threeparttable}
\usepackage{array}
\usepackage{booktabs}
\usepackage{bbding}
\usepackage{diagbox}
 \usepackage{graphicx}
 \usepackage{amsmath}
  \usepackage{amsfonts}
\usepackage[accsupp]{axessibility}  

\usepackage{hyperref}

\usepackage{orcidlink}

\begin{document}

\title{From Macro to Micro: Boosting micro-expression recognition via pre-training on macro-expression videos} 

\titlerunning{MA2MI}

\author{Hanting Li\inst{1} \and
Hongjing Niu\inst{1} \and
Feng Zhao\thanks{The corresponding author is Feng Zhao.}\inst{1}}

\authorrunning{H.~Li et al.}

\institute{University of Science and Techonology of China, Hefei, 230026, China \\
\email{\{ab828658,sasori\}@mail.ustc.edu.cn}\\
\email{\{fzhao956\}@ustc.edu.cn}}

\maketitle

\begin{abstract}
  Micro-expression recognition (MER) has drawn increasing attention in recent years due to its potential applications in intelligent medical and lie detection. However, the shortage of annotated data has been the major obstacle to further improve deep-learning based MER methods. Intuitively, utilizing sufficient macro-expression data to promote MER performance seems to be a feasible solution. However, the facial patterns of macro-expressions and micro-expressions are significantly different, which makes naive transfer learning methods difficult to deploy directly. To tacle this issue, we propose a generalized transfer learning paradigm, called \textbf{MA}cro-expression \textbf{TO} \textbf{MI}cro-expression (MA2MI). Under our paradigm, networks can learns the ability to represent subtle facial movement by reconstructing future frames. In addition, we also propose a two-branch micro-action network (MIACNet) to decouple facial position features and facial action features, which can help the network more accurately locate facial action locations. Extensive experiments on three popular MER benchmarks demonstrate the superiority of our method.
  \keywords{Micro-expression recognition \and Transfer learning \and Pre-training}
\end{abstract}

\section{Introduction}
\label{sec:intro}
Facial expression recognition (FER) is an essential way to analyze human emotions and is widely used in driver assistance system \cite{wilhelm2019dms}, healthcare aids \cite{muhammad2017health} and human-computer interaction \cite{abdat2011hci}. As a special type of facial expression, micro-expressions (MEs) reveal emotions that people try to hide, which makes micro-expression recognition (MER) become one of the major route for lie detection and mental health monitoring \cite{ekman2009lie}. However, the short duration of MEs and the small amplitude of facial movements make the recognition very difficult. In recent years, with the rapid development of deep learning technology, many MER methods based on neural networks have greatly improved the recognition performance \cite{ben2021mmew}.

\begin{figure}[tb]
  \centering
  \includegraphics[width=\linewidth]{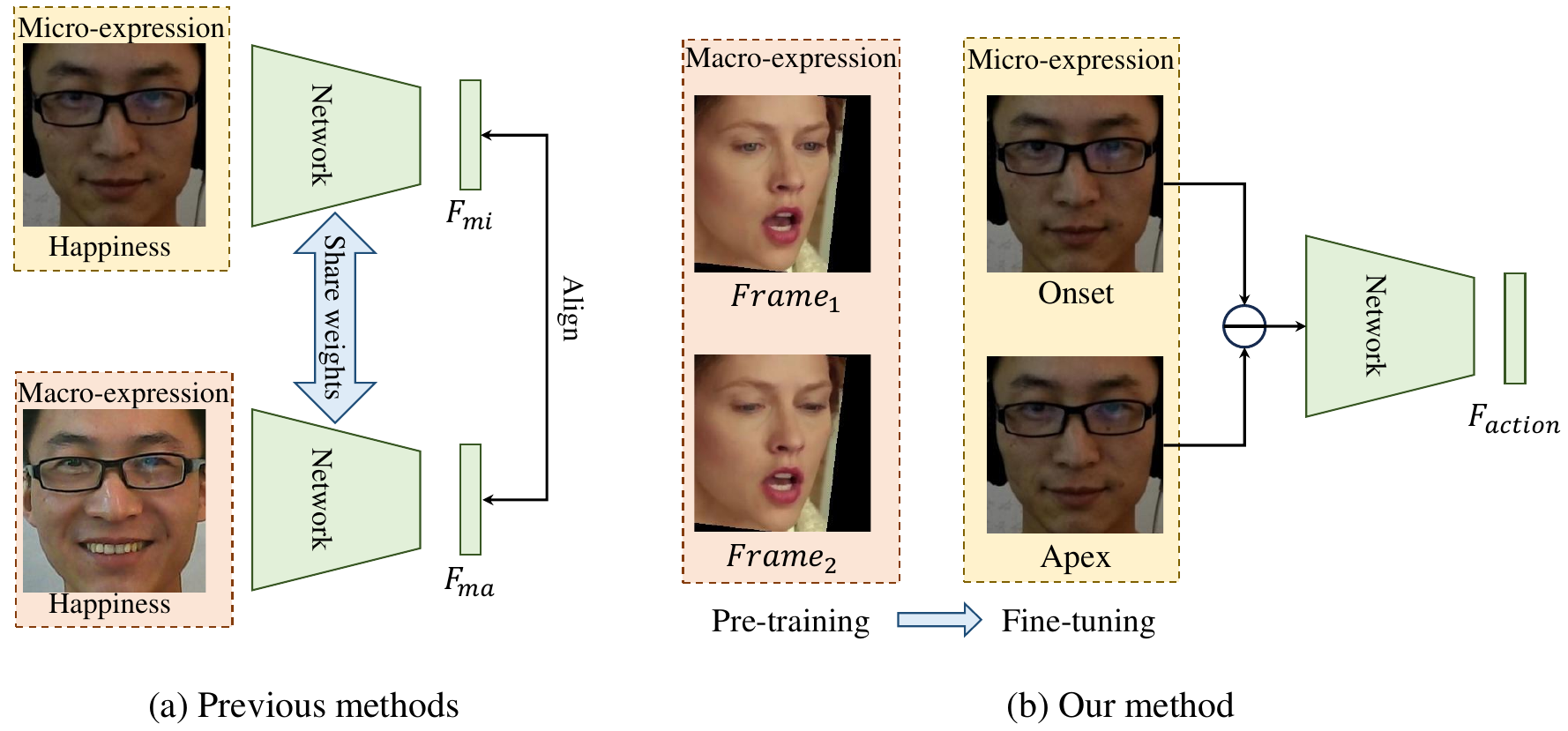}
  \caption{(a) Previous methods focus on finding common patterns (features) of macro-expressions and micro-expressions of the same category. $F_{mi}$ and $F_{ma}$ stand for features of micro- and macro-expressions, respectively. While (b) our method pre-trains the network on adjacent frames of macro-expression videos to obtain the ability to represent small facial actions.
  }
  \label{fig:1}
\end{figure}

As a well-established fact, deep-learning based techniques often rely on sufficient training data. When there is insufficient training data, the network can easily overfit the biased training data and affect the generalization performance \cite{ying2019overfit}. Due to the professional requirements for labeling micro-expression data, the amount of high-quality annotated data is very limited \cite{ben2022dall}. In contrast, the cost of annotating macro-expressions is much lower, and the amount of annontated data is more than 100 times that of micro-expressions. This has inspired many researchers to find common patterns of macro- and micro-expressions. Peng et al. follow the transfer learning paradigm to pre-train networks on macro-expression datasets and then fine-tune it on micro-expression datasets \cite{peng2018transfer}. Ben et al. build a benchmark that collects macro- and micro-expressions from same subjects, which provides support for research on the correlation between the two kinds of expressions \cite{ben2021mmew}. Xia et al. use a large amount of macro-expression data to guide the training of micro-expression recognition networks \cite{xia2020mame,xia2021mame}. Some researchers also try to map the micro-expressions embeddings into macro-expression embedding space through a translator, so that the classifier trained on macro-expression dataset can be adjusted and adapted to boost the classification performance
on the micro-expression dataset \cite{ben2022dall}. 

Although the above methods have gradually improved the performance of MER to a certain extent, they all rely on some ambiguous assumptions. That is, macro-expressions and micro-expressions of the same category have common visual action patterns \cite{liu2019mame}, which makes some algorithms require one-to-one correspondence between macro- and micro-expression categories. As shown in Fig.~\ref{fig:1}(a), previous methods focus on finding common patterns of two kinds of expressions by aligning their features. These constraints prevent them from being used on any macro-expression data, limiting the application scope of these methods. Since the core ability of MER method is to encode small facial actions between key frames (i.e., onset, apex, and offset frames) \cite{li2022survey}, we propose a generalized transfer learning paradigm, named \textbf{MA}cro-expression \textbf{TO} \textbf{MI}cro-expression (MA2MI). MA2MI acquire the ability to represent subtle facial movements through future frame reconstruction. In addition, we devise a two-branch micro-action network (MIACNet) to decouple facial position features and facial action features, which enables the network to locate facial movements of different subjects to specific facial areas. In this work, our contributions can be summarized as follow,

\begin{itemize}
    \item We propose a transfer learning paradigm that learns the ability to encode small facial movements by reconstructing future frames, named MA2MI. This training paradigm only require raw macro-expression data without annotations.
    \item We introduce a micro-action network that decouples facial position features and facial action features through two independent branches. 
    \item We conducted extensive experiments on three popular MER datasets and achieved state-of-the-art performance without cumbersome network structure design. In addition, the visualization results also demonstrate the rationality of the method design.
\end{itemize}

\section{Related Works}
\subsection{Micro-expression recognition}
As one of the key task in affective computing, micro-expression recognition methods have developed rapidly in the past decade \cite{li2022mersurvey}. Early research focused on designing stable hand-crafted features. Among them, local binary pattern (LBP) is one of the most commonly studied hand-crafted feature due to its strong ability to characterize local features \cite{ahonen2004face,pfister2011lbp,pfister2011lbp,zhao2007lbp}. In addition, optical flow-based features have also been widely studied because its ability to represent short-term motion information \cite{xu2017opticalflow,liu2015opticalflow,happy2017fuzzyopticalflow,li2019opticalflow}.

In recent years, with the rapid development of deep learning technology, deep-learning-based MER methods have gradually begun to show their advantages in generalization capabilities. Patel et al. pre-trained their network on macro-expression data to alleviate the challenges posed by insufficient training data for network training, and then select relevant features through evolutionary algorithms \cite{patel2016selective}. Gan et al. calculated the optical flow from the apex and onset frame, and then futher enhanced the optical flow feature through a convolutional neural network (CNN) \cite{gan2019off-apex}. Li et al. proposed a two-branch MER paradigm, which extract the facial position embeddings and muscle motion features from two independent networks \cite{li2022mmnet}. Specially, self-supervised learning methods are also used to pre-train networks by reconstructing images \cite{nguyen2023bert,fan2023selfme}. 

These methods have improved the performance of MER in various aspects. However, they are all suffered from lacking of annotated data and are easy to overfit the limited training data. Therefore, we proposed an transfer learning paradigm called MA2MI. By pre-training on a large amount of unlabeled macro-expression videos, we effectively alleviated the problem of lack of annotated micro-expression data and further boosted the MER performance.

\subsection{Transfer learning}
Transfer learning aims at improving the performance of target learners on target domains by transferring the knowledge
contained in different but related source domains \cite{zhuang2020transfer}. In this way, the reliance on large amounts of target domain data for building target learners can be reduced. In general, transfer learning methods can be divided into two categories according to the discrepancy between source and target domain, i.e., homogeneous \cite{huang2006correcting} and heterogeneous \cite{day2017survey} transfer learning. When the source and the target domain have same feature spaces and label spaces, the task can be taken as homogeneous transfer learning, otherwise it belongs to heterogeneous transfer learning \cite{tang2023review,weiss2016survey}.

For MER task with very limited annotated data, transferring knowledge from other source is an effective way to improve MER performance. Jia et al. proposed a macro-to-micro transformation model which enables to transfer macro-expression learning to micro-expression \cite{jia2018SVD}. Zhu et al. leveraged rich speech data to enhance MER by transferring learning from the speech to the MER \cite{zhu2018coupled}. Zong et al. devised a transductive transfer regression model to bridge the feature distribution gap between the source and target domains by learning a joint regression model \cite{zong2019da}. Sun et al. utilized knowledge from action unit under a knowledge distillation paradigm \cite{sun2020KD}. Peng et al. and Razak et al. directly took advantage of macro-expression data by pre-training networks on macro-expression datasets \cite{peng2018transfer,ab2023lightweightmame}. The above works focus on expanding the training data size, but ignores that the core ability of MER is to capture small facial actions. To address this issue, our proposed transfer learning paradigm learns the ability to encode subtle facial movements from macro-expression videos, which better adapts to the target domain task (i.e., MER).

\subsection{Macro-expression boosted micro-expression recognition}
Recently, researchers begin to use large amounts of macro-expression data to improve MER performance. For MER task with very limited annotated data, macro-expressions, which are also facial expressions, seem to be a perfect data source to improve MER performance. Liu et al. magnificated micro-expression while reducing macro-expressions, thereby narrow the gap between these two kinds of facial expressions \cite{liu2019mame}. Peng et al. and Razak et al. pre-trained networks on macro-expression recognition datasets, which requires that macro-expression and micro-expression data have the same label space \cite{peng2018transfer,ab2023lightweightmame}. Xia et al. introduced two expression identity disentangle network, named MicroNet and MacroNet, as the feature extractors. MacroNet is then fixed and used to guide the fine-tuning of MicroNet from both label and feature space \cite{xia2020mame}. Ben et al. proposed an active learning method of making uses of the unlabeled data in the training dataset, meanwhile aligns these data with the data in macro-expression domain, and uses the classifier in macro-expression domain to predict and recognize micro-expressions \cite{ben2022dall}. 

Since macro- and micro-expressions have similar label space (e.g., happiness, sadness, anger, and surprise), previous methods naturally assume they share a common feature space \cite{peng2018transfer,ab2023lightweightmame}. However, as shown in Fig.~\ref{fig:1}(a), the difference between macro- and micro-expression data is very significant, which is mainly reflected in the intensity of the apex frame \cite{allaert2022mame_relation}. In addition, the available macro- and micro-expression data may not have aligned label space. Therefore, in this work, we do not assume that macro- and micro-expressions share the same feature space and label space, which means that our MA2MI has a wider application scope.

\section{Method}
In this section, we first introduce MA2MI, an transfer learning framework. Then the proposed two-branch micro-action network structure is detailed.

\subsection{MIACNet: Decouple Facial Position and Action Features}
\begin{figure}[tb]
  \centering
  \includegraphics[width=0.8\linewidth]{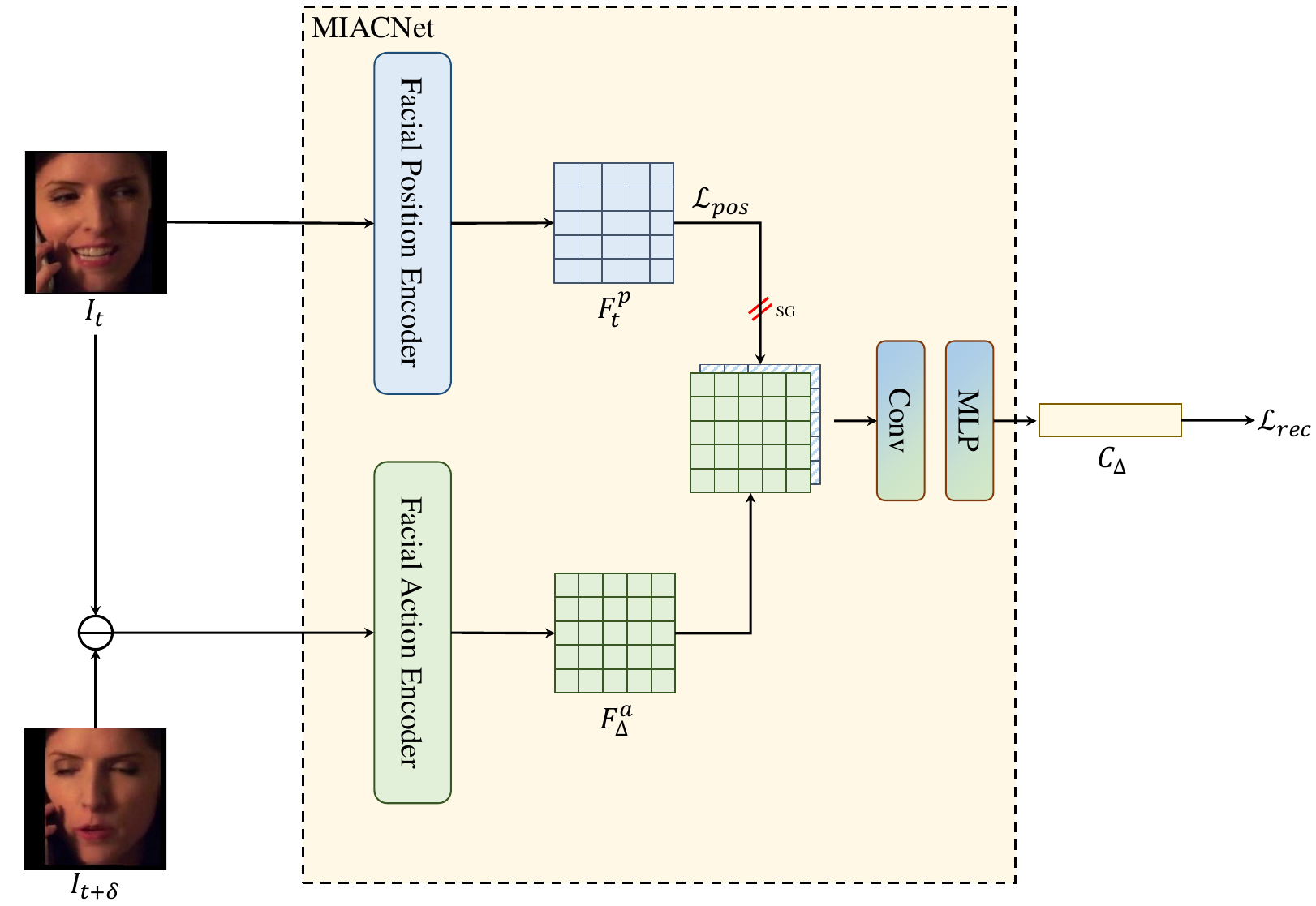}
  \caption{MIACNet for encoding subtle facial actions between $I_t$ and $I_{t+\delta}$.
  }
  \label{fig:miacnet}
\end{figure}

There are two key aspects to recognize micro-expressions, which are the location where facial actions occur and the facial action patterns \cite{li2022mmnet}. As shown in Fig.~\ref{fig:miacnet}, we propose micro-action network (MIACNet) to extract subtle facial actions between temporal neighbor frames. In order to learn facial position and action features separately without interfering with each other, MIACNet consists of two independent encoders (i.e., facial position encoder and facial action encoder). We directly utilize ResNet18 \cite{he2016resnet} as the encoder to demonstrate the generality of our approach.

\begin{figure}[tb]
  \centering
  \includegraphics[width=0.8\linewidth]{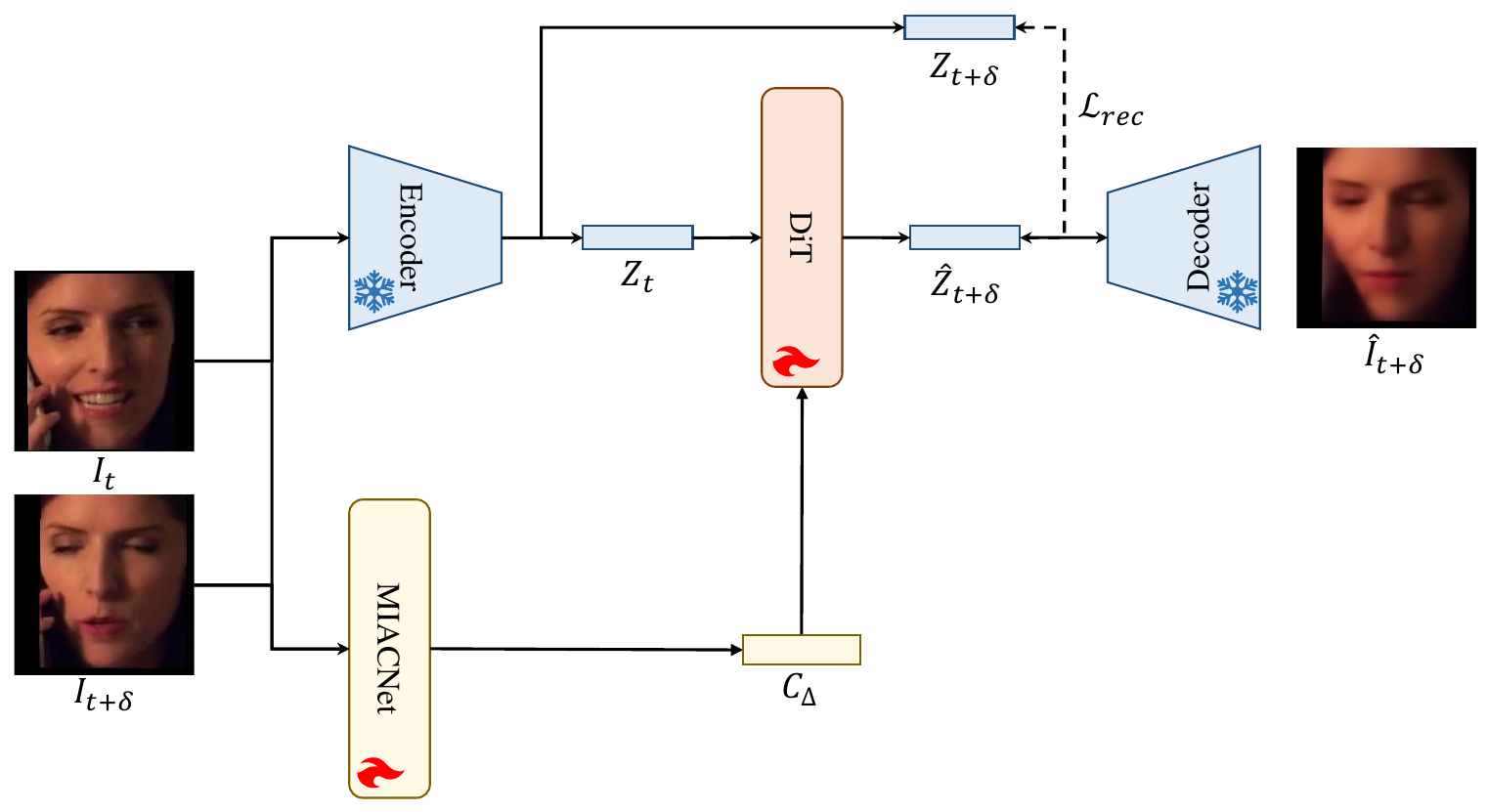}
  \caption{The pre-training process on macro-expression data.
  }
  \label{fig:pretrain}
\end{figure}

As shown in Fig.~\ref{fig:miacnet}, $I_t$ and $I_{t+\delta}$ are sampled from facial expression videos. $t$ is a random initial time. In order to obtain short-term facial actions, the value of $\delta$ is a small positive integer, which represents the sample interval. For facial action branch, the difference between $I_t$ and $I_{t+\delta}$ is taken as the input, which can be formulated as,
\begin{equation}\label{eq:1}
F^a_{\Delta}=\rm{E_a}\it(I_{t+\delta}-I_{t}).
\end{equation}
\noindent Where $E_a$ stands for the facial action encoder. This encoder is trained by 
minimizing the reconstruction loss $\mathcal{L}_{rec}$ as shown in Fig.~\ref{fig:pretrain}. The details will be detailed in the following chapter.

As for the facial position encoder, the facial position feature $F_t^p$ is supposed to distinguish between different facial areas, so that $F_t^p$ can be used to pinpoint the facial position where the micro-action occurs at $t$. In order to avoid the entanglement of action and position features, we designed $L_{pos}$ for training the facial position encoder.

Facial position features need to have three characteristics, according to which $\mathcal{L}_{pos}$ can be divided into three parts. First, the premise that position features can locate faces is that the features corresponding to different areas of the face are different, so the first part can be defined as,

\begin{equation}\label{eq:2}
\mathcal{L}_{1} = \frac{\sum_{i=1}^{HW}\sum_{j=1,j\neq i}^{HW}\langle F^p(i), F^p(j) \rangle}{HW(HW-1)}.
\end{equation}
\noindent Where $H$ and $W$ stand for the width and height of the facial position feature $F^p$, and $F^p(i)$ represent the i-$th$ position at the spatial plane. $\langle\cdot, \cdot\rangle$ is the cosine similarity between two vectors with the same size. This loss ensures the difference in features in different facial areas and facilitates subsequent positioning of sub-actions.

The second part of the $\mathcal{L}_{pos}$ is to ensure cross-face consistency in each facial areas (e.g., left mouth corner and Right eyebrow). This is because the facial position features need to remain unified across different faces and not be affected by irrelevant information such as identity. This part can be mathematically expressed as follows,

\begin{equation}\label{eq:3}
\mathcal{L}_{2} = \frac{\sum_{i=1}^{HW}\sum_{j=1, j\neq j_i^*}^{HW}\langle F_1^p(i), F_2^p(j) \rangle}{HW(HW-1)}-\frac{\sum_{i=1}^{HW}\max\limits_{j}\langle F_1^p(i), F_2^p(j) \rangle}{HW},
\end{equation}

\noindent with

\begin{equation}\label{eq:4}
j_i^* = \underset{j\in\{1,2,\cdot\cdot\cdot,HW\}}{\arg\max}\langle F_1^p(i), F_2^p(j) \rangle.
\end{equation}

\noindent Where $F_1^p$ and $F_2^p$ denote the facial position feature of two different facial images. $j_i^*$ stands for the represents the spatial index of $F_2^p$ that uniquely corresponds to $F_1^p(i)$. By minimizing $\mathcal{L}_{2}$, different face position features can be matched one-to-one at spatial level to ensure their consistency across faces.

The last part of the $\mathcal{L}_{pos}$ aims to make the input $I$ and $F^p$ consistent with the spatial transformation (e.g., rotation and translation). Therefore, 
$\mathcal{L}_{3}$ can be calculated by,
\begin{equation}\label{eq:5}
\mathcal{L}_{3} =  \lVert \rm E_p \it (\tau(I))- \tau(\rm E_p \it(I)) \rVert_2,
\end{equation}
\noindent where $\rm E_p$ and $I$ stand for the facial position encoder and its input, respectively. $\tau$ is random spatial augmentation and $\lVert \cdot \rVert_2$ represents 2-Norm. $\mathcal{L}_{3}$ ensures the spatial sensitivity of $F^p$.

The $\mathcal{L}_{pos}$ used to train the facial position branch can be written as,
\begin{equation}\label{eq:6}
\mathcal{L}_{pos} =  \mathcal{L}_{1}+\mathcal{L}_{2}+\mathcal{L}_{3}
\end{equation}

\subsection{MA2MI: An generalized transfer learning paradigm for MER}
In order to obtain a wider applicability, our transfer learning paradigm does not make any assumptions about the feature and label space of macro-expression and micro-expression data. We focus on obtaining the core capabilities required to recognize micro-expressions from a large amount of macro-expression data, that is, the ability to encode small facial actions. Transfer learning is generally divided into two steps: pre-training on source domain and fine-tuning on target domain. We will detail these two parts respectively.

\subsubsection{Pre-training on Macro-expression data}
Macro-and micro-expressions have significant differences in visual patterns, which undoubtedly hinders the development of corresponding transfer learning methods. Micro-expressions are tiny facial movements that occur in a very short period of time \cite{ekman2009lie}. As shown in Fig.~\ref{fig:pretrain}, according to this characteristic, we design the pre-training process based on latent-space reconstruction of near-future frame. 

For each macro-expression video, we sample two frames $I_t$ and $I_{t+\delta}$. The network is divided into two branches, which are the reconstruction branch for conditional generation and the conditional branch for facial micro-action encoding. For encoding the facial micro-actions, the proposed MICANet which can encodes the facial actions between $I_t$ and $I_{t+\delta}$ into a condition vector $C_\Delta$. For the reconstruction branch, we generally follow DiT \cite{peebles2023DIT} to complete the conditional generation part in latent space since reconstruction in high-resolution pixel space can be computationally
prohibitive. Therefore, we directly use the  autoencoder in \cite{peebles2023DIT} to compresse $I_t$ and $I_{t+\delta}$ into smaller spatial representations $Z_t$ and $Z_{t+\delta}$, which can be formally defined as,

\begin{equation}\label{eq:7}
\begin{aligned}
&Z_t = \rm E_{ae} \it (I_t), \\
&Z_{t+\delta} = \rm E_{ae} \it (I_{t+\delta}).
\end{aligned}
\end{equation}

\noindent Where $\rm E_{ae}$ denote the encoder of the autoencoder. Then $Z_t$ and $C_\Delta$ are fed into DiT to predict the latent embedding of $I_{t+\delta}$, 
\begin{equation}\label{eq:8}
\hat{Z}_{t+\delta} = \rm DiT \it(Z_t,C_\Delta).
\end{equation}
\noindent And the reconstruction loss can then be formulated as,
\begin{equation}\label{eq:9}
\mathcal{L}_{rec} = \lVert \hat{Z}_{t+\delta}-Z_{t+\delta} \rVert_1.
\end{equation}
\noindent Where $\lVert \cdot \rVert_1$ represents the 1-norm. The overall loss of the pre-training process is defined as,
\begin{equation}\label{eq:10}
\mathcal{L}_{pre} = \mathcal{L}_{rec}+\mathcal{L}_{pos}.
\end{equation}

\subsubsection{Fine-tuning on Micro-expression data}
In this stage, we use a small amount of annotated micro-expression data to further fine-tune MIACNet to adapt to the MER task. Since one of the main MER approaches is recognizing the small facial movements between key frames (i.e., onset, apex, and offset frames) \cite{li2022survey}, we only need to replace $I_t$ and $I_{t+\delta}$ with onset and apex frame so that MIACNet can encode the micro-expression into $C_\Delta$. $C_\Delta$ is then projected to a N-dimension vector through a single fully connected (FC) layer for N-class micro-expression recognition. Thanks to the ability to encode small facial movements acquired during the pre-training process, MIACNet can achieve advanced performance by fine-tuning on small-scale annotated micro-expression data.

\section{Experiments}
To verify the effectiveness of MA2MI, we pre-train our model on macro-expression datasets (i.e., DFEW \cite{jiang2020dfew}, FERV39K \cite{wang2022ferv39k}, and AFEW \cite{dhall2012afew}). Then the model is fine-tuned on three micro-expression datasets respectively, including CASME II \cite{yan2014casmeii}, SAMM \cite{davison2016samm}, and MMEW \cite{ben2021mmew}. We first introduce the used datasets, evaluation protocols, and present the implementation details. Then extensive ablation studies are conducted to demonstrate the effectiveness of our method.

\subsection{Datasets}
\noindent\textbf{Macro-expression Dataset}

\noindent\textbf{DFEW} \cite{jiang2020dfew} consists of over 16,000 video clips from thousands of movies. Each video clip is individually annotated by ten independent individuals under professional guidance and assigned to one of seven basic expressions (i.e., happiness, sadness, neutral, anger, surprise, disgust, and fear). Since the proposed MA2MI is a reconstruction-based pre-training method, our method does not rely on manual annotation.

\noindent\textbf{FERV39K} \cite{wang2022ferv39k} is currently the largest in-the-wild DFER dataset and contains 38,935 video sequences collected from 4 scenarios, which can be further divided into 22 fine-grained scenarios, such as crime, daily life, speech, and war. Each clip is annotated by 30 individual annotators and assigned to one of the seven basic expressions as DFEW.

\noindent\textbf{AFEW} \cite{dhall2012afew} served as an evaluation platform for the annual EmotiW from 2013 to 2019 that contains 1,809 video clips collected from movies. All the clips are split into training set (773 video clips), validation set (383 video clips), and testing set (653 video clips).

\noindent\textbf{Micro-expression Datasets}

\noindent\textbf{CASME II} \cite{yan2014casmeii} collects 256 micro-expression videos sourced from 26 subjects, captured at 200 FPS. The manual annotation include onset/apex/offset frames, action units, and emotions. We only use the samples of happiness, disgust, repression, surprise, and others for 5-class MER.

\noindent\textbf{SAMM} \cite{davison2016samm} consist of 159 ME clips from 32 participants of 13 different ethnicities at 200 FPS. Onset/apex/offset frames, action units, and emotions are also carefully annotated. Five prototypical expressions (happiness, anger, contempt, surprise, and others) are utilized for experiments.

\noindent\textbf{MMEW} \cite{ben2021mmew} contains both macro- and micro-expressions sampled from the same subjects. Specifically, it consists of 300 micro-expressions and 900 macro-expressions, which are collected at 90 FPS. Consistent with the official setting, we use samples of happiness, disgust, surprise, sadness, anger, and fear for training and testing.

\subsection{Evaluation Protocols}
For CASME II and SAMM datasets, leave-one-subject-out (LOSO) cross-validation is employed as the evaluation protocol. Under this protocol, each subject is taken as the test set in turn and the rest is taken as the training set. Consistent with the official protocol in \cite{ben2021mmew}, we adopt the five-fold cross-validation protocol. Specially, all samples are randomly split into five subsets according to ``subject independent'' criterion. For CASME II and SAMM, the accuracy and the unweighted F1-score (UF1) are used for evaluation. UF1 can be calculated by,
\begin{equation}\label{eq:11}
 UF1 = \frac{1}{N_c}\sum_{i=1}^{N_c} F1_i,
\end{equation}
\noindent where
\begin{equation}\label{eq:12}
 F1_i = \frac{2TP_i}{2TP_i+FN_i+FP_i}.
\end{equation}
\noindent $F1_i$ is the F1-score of the $i$-th class and $N_c$ represents the number of the class. $TP_i$, $FN_i$, and $FP_i$ are the number of true positive, false negative, and false positive samples respectively. While for MMEW, only the accuracy is reported as the metric which is also consistent with the official setting in \cite{ben2021mmew}.

\subsection{Implementation Details}
In all the experiments, all the video frames are resized to $256$$\times$$256$ for training and testing. For the pre-training process, we use AdamW optimizer \cite{loshchilov2017adamw} to optimize MIACNet and DiT-B \cite{peebles2023DIT} with a batch size of 32. DFEW dataset is taken as the default macro-expression dataset. The learning rate is initialized to 0.0004, decreased at an exponential rate in 80 epochs. The sample interval $\delta$ belongs to $\{3,4,5,6,7,8\}$ by default. At the fine-tuning stage, we also use AdamW optimizer to fine-tune MIACNet with a zero-initialized FC layer on MER datasets with a batch size of 16 for 80 epochs. The learning rate is set to 0.0004 and the weight decay is 0.1. The random cropping, horizontal flipping, and random rotation are employed to avoid over-fitting. All the experiments are conducted on a single NVIDIA RTX 3090 card with PyTorch toolbox \cite{paszke2019pytorch}.

\begin{table}[t]
\begin{center}
\caption{Evaluation of different pre-training method. MAER stands for macro-expression recognition task. ``w/o'' means without pre-training on macro-expression datasets. $^\dagger$ denotes reconstruction of $I_{t+\delta}$ in pixel space. N/A not applicable. The best results are highlighted in bold.
}
\begin{tabular}{c|c|cc|cc|c}
\toprule
\multirow{2}{*}{Pre-training Setting}&\multirow{2}{*}{Annotation}&\multicolumn{2}{c|}{CASME II }&\multicolumn{2}{c|}{SAMM }&\multicolumn{1}{c}{MMEW}\cr
    \cmidrule(lr){3-7}&& Acc (\%)& UF1& Acc (\%)& UF1& Acc (\%)\cr
    
\midrule
w/o (baseline)&N/A&83.94&0.8073&77.21&0.6740&68.80\cr
\midrule
MAER&\CheckmarkBold&83.53&0.8166&78.68&0.7214&72.22\cr

\midrule

MA2MI$^\dagger$& \XSolidBrush&87.55&0.8732&80.88&0.7481&74.36\cr
MA2MI& \XSolidBrush&\textbf{89.16}&\textbf{0.8882}&\textbf{83.82}&\textbf{0.7893}&\textbf{75.21}\cr
\bottomrule
\end{tabular}
\label{tab:pre}
\end{center}
\end{table}

\subsection{Ablation Studies}
\textbf{Evaluation of Different Pre-training Strategies:} Previous transfer learning methods were mostly based on macro-expression recognition tasks to find common patterns between macro- and micro-expressions of a same category, thereby improving MER performance \cite{liu2019mame,ab2023lightweightmame}. Such methods rely on high-quality annotation and sometimes even require alignment label spaces (i.e., one-to-one correspondence between expression categories). In addition, the main paradigm based on key frames in MER is also significantly different from the mainstream methods of macro-expression recognition. These limitations and differences greatly affect the scalability of the method. 

We compare classic pre-training methods based on macro expression recognition tasks in Table~\ref{tab:pre}. For fair comparison, all networks adopt the proposed MIACNet. Pre-training based on macro-expression recognition (MAER) task is implement by training the network through cross entropy loss. The results show that the improvement obtained through MAER is very limited, and there is even no improvement on CASME II, which is mainly due to the large difference in the visual pattern of two kinds of facial expressions. In comparison, MA2MI significantly exceeds the performance of the baseline on all datasets. Besides, the performance of conduct MA2MI in high-resolution pixel space is also compared. Although the performance is equivalent to that of MA2MI in latent space, it 
can be computationally prohibitive.

\begin{table}[t]
\begin{center}
\caption{Evaluation of different fine-tuning method. Reconstruction indicates whether to retain the reconstruction part of the pre-training process in the fine-tuning stage. FPE and FAE stand for the facial position and facial action encoder of MIACNet, respectively. The best results are highlighted in bold.
}
\begin{tabular}{c|cc|cc|cc|c}
\toprule
\multirow{2}{*}{Reconstruction}&\multicolumn{2}{c|}{Branches}&\multicolumn{2}{c|}{CASME II }&\multicolumn{2}{c|}{SAMM }&\multicolumn{1}{c}{MMEW}\cr
    \cmidrule(lr){2-8}&FPE&FAE& Acc (\%)& UF1& Acc (\%)& UF1& Acc (\%)\cr
    
\midrule
\CheckmarkBold&\CheckmarkBold&\XSolidBrush&83.94&0.8073&80.88&0.7304&72.65\cr
\CheckmarkBold&\XSolidBrush&\CheckmarkBold&85.54&0.7941&80.88&0.7637&73.08\cr
\CheckmarkBold&\CheckmarkBold&\CheckmarkBold&88.76&0.8795&81.62&0.7384&73.93\cr

\midrule
\XSolidBrush&\CheckmarkBold&\XSolidBrush&84.34&0.8311&81.62&0.7411&73.93\cr
\XSolidBrush&\XSolidBrush&\CheckmarkBold&87.95&0.8486&82.35&0.7640&74.78\cr
\XSolidBrush&\CheckmarkBold&\CheckmarkBold&\textbf{89.16}&\textbf{0.8882}&\textbf{83.82}&\textbf{0.7893}&\textbf{75.21}\cr
\bottomrule
\end{tabular}\vspace{-0.3cm}
\label{tab:fine}
\end{center}
\end{table}
\begin{table}[t]
\begin{center}
\caption{Evaluation of pre-training on different macro-expression datasets. The best results are highlighted in bold.}
\begin{tabular}{c|cc|cc|c}
\toprule
\multirow{2}{*}{\diagbox{Source Dataset}{Target Dataset}}&\multicolumn{2}{c|}{CASME II }&\multicolumn{2}{c|}{SAMM }&\multicolumn{1}{c}{MMEW}\cr
    \cmidrule(lr){2-6}& Acc (\%)& UF1& Acc (\%)& UF1& Acc (\%)\cr
    
\midrule
None&83.94&0.8073&77.21&0.6740&68.80\cr
\midrule
AFEW \cite{dhall2012afew}&85.14&0.8362&80.15&0.7146&72.22\cr
FERV39K \cite{wang2022ferv39k}&\textbf{89.96}&\textbf{0.8964}&81.62&0.7640&74.36\cr
DFEW \cite{jiang2020dfew}&89.16&0.8882&\textbf{83.82}&\textbf{0.7893}&\textbf{75.21}\cr
\bottomrule
\end{tabular}\vspace{-0.5cm}
\label{tab:madataset}
\end{center}
\end{table}

\noindent\textbf{Evaluation of Different Fine-tuning Strategies:} In transfer learning, fine-tuning on target domain is equally important as pre-training on source domain. Therefore, we evaluate the impact of different fine-tuning strategies in Table~\ref{tab:fine}. First we investigated whether the reconstruction part in the pre-training phase should be maintained during the fine-tuning phase. The results show that introducing reconstruction tasks to assist in the fine-tuning process hinders further performance improvement. This is because not all facial movements are related to micro-expressions, which means that maintaining the reconstruction task may learn irrelevant facial movements (e.g., blink) and thus affect the classification. In addition, we also study whether all parameters should be tuned. Consistent with consensual experience, full-parameter fine-tuning can achieve the best performance. Fine-tuning the parameters of facial action encoder will improve performance more than only fine-tuning facial position encoder.

\noindent\textbf{Pre-training on Different Macro-expression Datasets} To verify the generality of our method, we pre-train networks on three different macro-expression datasets respectively. As shown in Table~\ref{tab:madataset}, pre-training on different macro-expression datasets can boost MER performance. Specially, training on the larger macro-expression dataset (i.e., DFEW and FERV39K) can obtain better results than training on small-scale one (i.e., AFEW).

\begin{figure}[t]
\centering  
	\subfloat[The impact of $\delta$ on accuracy.]{
		\includegraphics[scale=0.26]{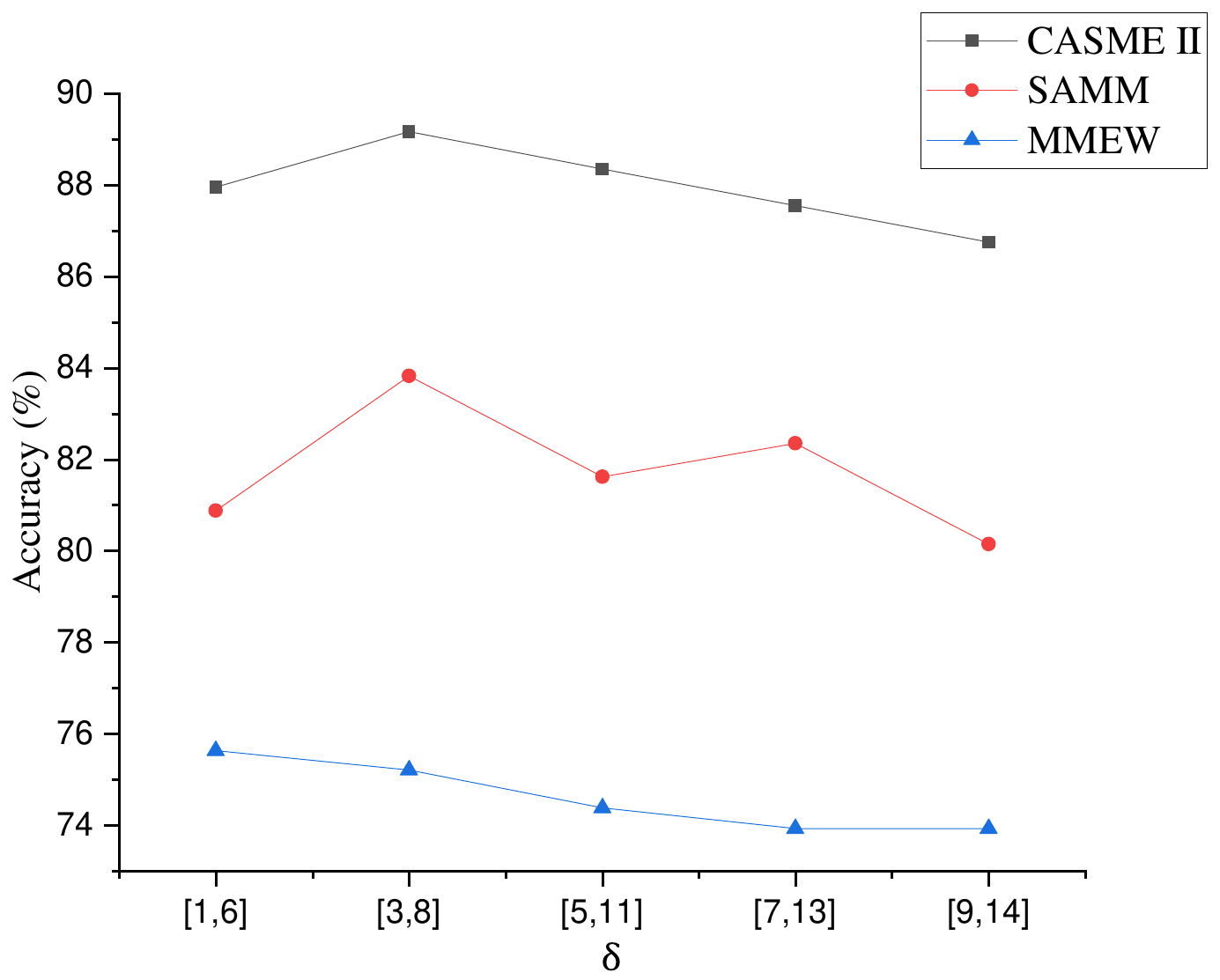}}
\subfloat[The impact of $\delta$ on UF1.]{
		\includegraphics[scale=0.26]{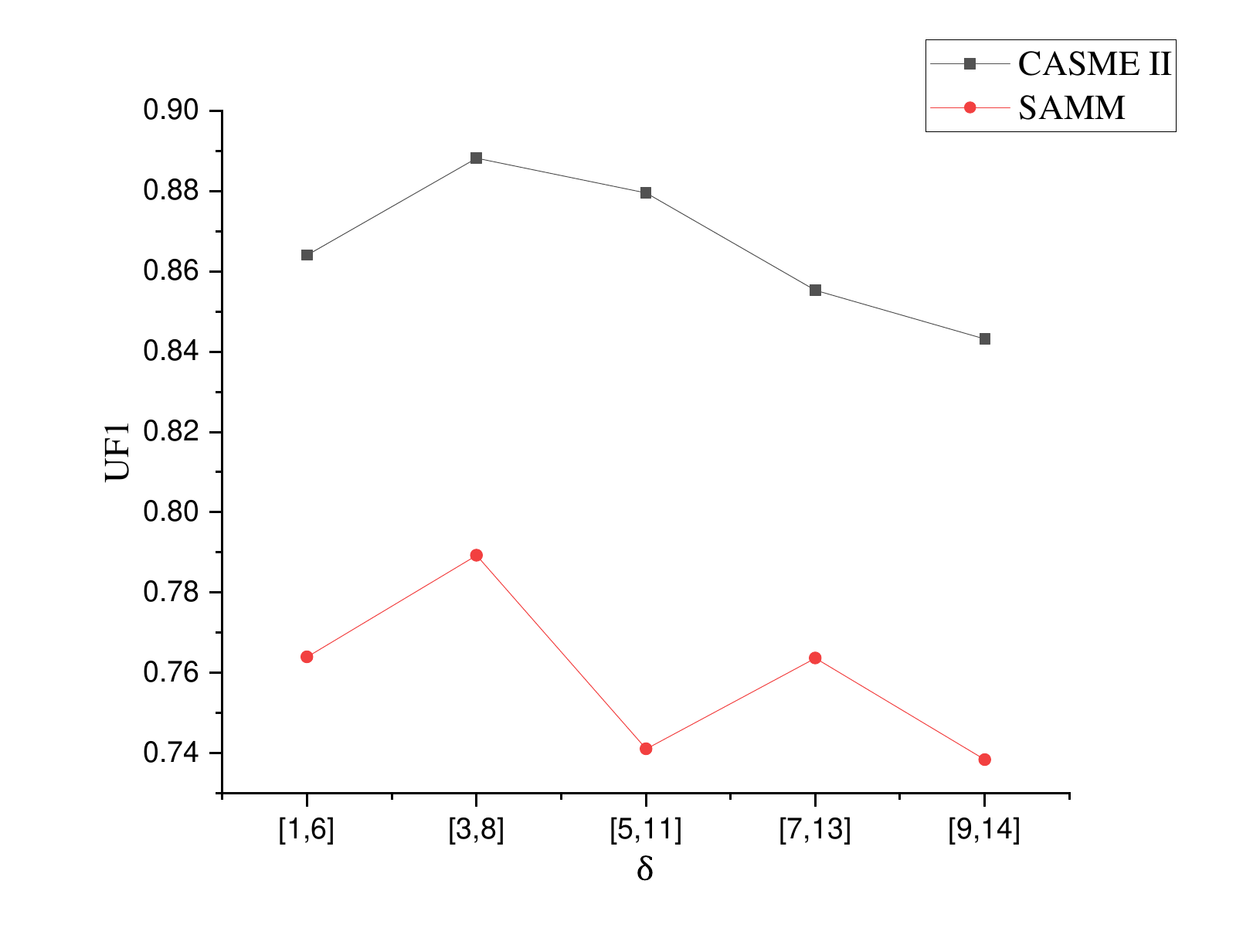}}
 \caption{The impact of $\delta$ on the performance of the proposed MA2MI on three datasets. The horizontal axis indicates that $\delta$ is any integer belongs to $[a,b]$.}
  \label{fig:delta}
\end{figure} 

\begin{table}[t]
\begin{center}
\caption{Evaluation of facial position encoder and $\mathcal{L}_{pos}$. The second setting means that both encoders are optimized through $\mathcal{L}_{rec}$ only. The best results are highlighted in bold.
}\vspace{-0.3cm}
\begin{tabular}{cc|cc|cc|c}
\toprule
\multirow{2}{*}{$\mathcal{L}_{pos}$}&\multirow{2}{*}{FPE}&\multicolumn{2}{c|}{CASME II }&\multicolumn{2}{c|}{SAMM }&\multicolumn{1}{c}{MMEW}\cr
    \cmidrule(lr){3-7}&& Acc (\%)& UF1& Acc (\%)& UF1& Acc (\%)\cr
    
\midrule
\XSolidBrush&\XSolidBrush&84.74&0.8372&77.94&0.7279&70.94\cr
\XSolidBrush&\CheckmarkBold&88.76&0.8795&81.62&0.7549&74.35\cr
\midrule
\CheckmarkBold&\CheckmarkBold&\textbf{89.16}&\textbf{0.8882}&\textbf{83.82}&\textbf{0.7893}&\textbf{75.21}\cr
\bottomrule
\end{tabular}\vspace{-0.3cm}
\label{tab:fpe}
\end{center}
\end{table}
\noindent\textbf{Evaluation of Different Sampling Interval} $\delta$is an important hyperparameter in our method, which represents the sampling interval between frame pairs. An excessively large sampling interval is not conducive for MIACNet to obtain ability to encode short-term subtle facial actions, while an excessively small sampling interval can easily cause the network to converge to a trivial solution (i.e., directly taking $I_t$ as the prediction of $I_{t+\delta}$). In Fig.~\ref{fig:delta}, we investigated the impact of different sampling intervals on the final performance. In the horizontal axis coordinate, $[a, b]$ represents $\delta$$\in$$Z^+$ is randomly sampled from $a$ to $b$. The results show that $[3,8]$ is a suitable sampling interval for all datasets, which will also be used as the default sampling interval in this work.

\noindent\textbf{Effectiveness of the Facial Position Encoder in MIACNet} In this work, the role of $C_\Delta$ is to represent the subtle actions between $I_t$ and $I_{t+\delta}$. This seems to mean that the face position branch with $I_t$ as input is not necessary. In Table~\ref{tab:fpe}, we study whether we should introduce facial position encoder and whether the two encoders should be trained independently. The results shows that introducing the facial position encoder can significantly improve the performance in terms of accuracy and UF1. Since $\mathcal{L}_{pos}$ decouples facial position features from identity information, the performance can be further improved.

\begin{table}[t]\small
    \caption{Comparison with State-of-the-Arts on CASME II and SAMM.}\vspace{-0.3cm}
    \begin{subtable}[h]{0.5\linewidth}
        \centering
        \caption{Comparison on CASME II.}
        \begin{tabular}{c|cc}
        \toprule  
            Method  & Accuracy (\%) &UF1  \\
			\midrule
		
            DSSN \cite{khor2019DSSN}&71.19&0.7297\\
            TSCNN \cite{song2019TSCNN}&80.97&0.8070\\
            Dynamic \cite{sun2020KD}&72.61&0.6700\\
            Graph-TCN \cite{lei2020Graph-TCN}&73.98&0.7246\\
            SMA-STN \cite{liu2020SMA-STN}&82.59&0.7946\\
            AU-GCN \cite{lei2021micro}&74.27&0.7047\\
            GEME \cite{nie2021geme}&75.20&0.7354\\
            MERSiamC3D \cite{zhao2021MERSiamC3D}&81.89&0.8300\\
            MMNet \cite{li2022mmnet} & 88.35&0.8676 \\
			\midrule
            MA2MI (Ours) & \textbf{89.16}&\textbf{0.8882}\\
            
			\bottomrule
        \end{tabular}
       \label{subtab:casme}
    \end{subtable}
    \hfill
    \begin{subtable}[h]{0.5\linewidth}
        \centering
        \caption{Comparison on SAMM.}
        \begin{tabular}{c|cc}
			\toprule  
			Method   & Accuracy (\%) &UF1  \\
			\midrule
            DSSN \cite{khor2019DSSN}&57.35&0.4644\\
            Graph-TCN \cite{lei2020Graph-TCN}&75.00&0.6985\\
            SMA-STN \cite{liu2020SMA-STN}&77.20&0.7033\\
            AU-GCN \cite{lei2021micro}&74.26&0.7045\\
            GEME \cite{nie2021geme}&55.38&0.4538\\
            MERSiamC3D \cite{zhao2021MERSiamC3D}&68.75&0.6400\\
            MMNet \cite{li2022mmnet} & 80.14&0.7291 \\
            \midrule
            MA2MI (Ours) & \textbf{83.82}&\textbf{0.7893}\\
            
			\bottomrule
		\end{tabular}
       \label{subtab:samm}
    \end{subtable}
    
     \label{tab:sota1}
\end{table}

\begin{table}[t]
\caption{Comparison with state-of-the-arts on MMEW.}\vspace{-0.5cm}
	\begin{center}
		\begin{tabular}{c|c}
			\toprule  
			Method   & Accuracy (\%)   \\
			\midrule
            LBP-TOP \cite{zhao2007dynamic}&38.90\\
            KGSL \cite{zong2018kgsl}&56.90\\
            MDMO \cite{liu2015MDMO}&65.70\\
            TLCNN \cite{wang2018tlcnn}&69.40\\
            Sparse Transformer\cite{zhu2022sparse}&70.59\\
            LD-FMERN \cite{ni2023LD-FMERN}&71.70\\
            \midrule
            MA2MI (Ours) & \textbf{75.21}\\
            
			\bottomrule
		\end{tabular}\label{tab:sota2}\vspace{-0.7cm}
	\end{center}
\end{table}

\subsection{Comparison with State-of-the-Arts} 
We compared our MA2MI with existing state-of-the-art methods on three popular MER benchmarks in Table \ref{tab:sota1}/\ref{tab:sota2}. The results of our method exceed previous methods on three datasets, which demonstrate the effectiveness of MA2MI. It should be note that MIACNet is not specially designed and consists of only two classic ResNet18 \cite{he2016resnet}. Therefore, the gain of the MA2MI comes entirely from the proposed transfer learning paradigm.

\begin{figure}[t]
	\centering
	\begin{minipage}{0.45\linewidth}
		\centering
		\includegraphics[width=0.95\linewidth]{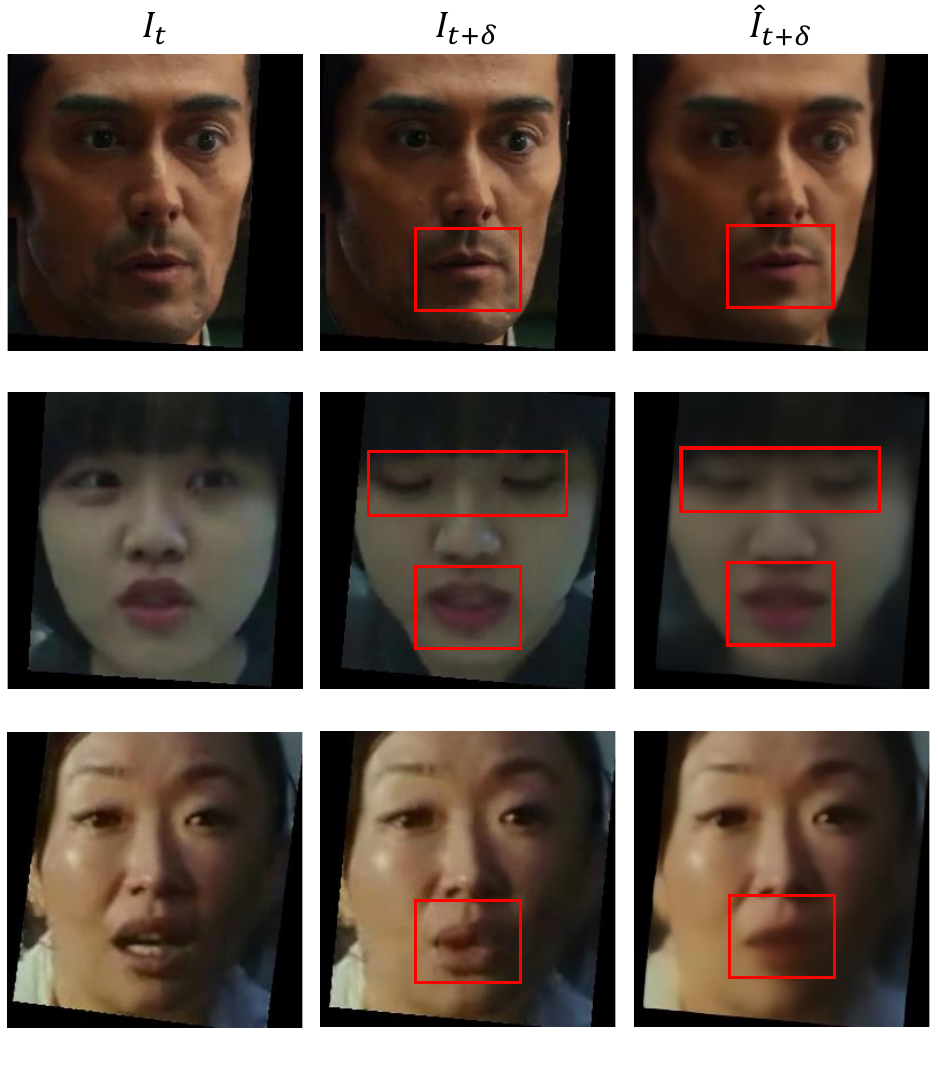}
		\caption{Reconstruction results on DFEW. $\hat{I}_{t+\delta}$ is reconstructed from $I_t$ and $C_\Delta$.}
		\label{fig:rec}
	\end{minipage}
	\begin{minipage}{0.45\linewidth}
		\centering
		\includegraphics[width=0.95\linewidth]{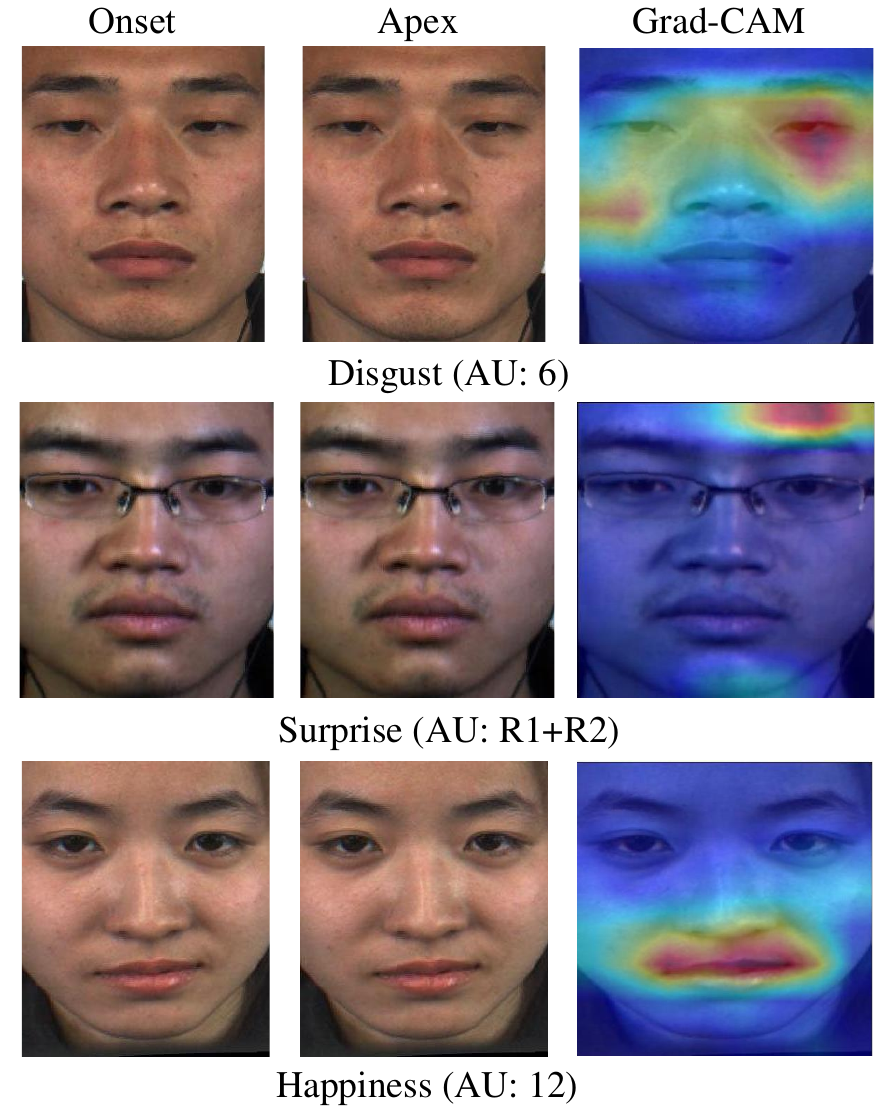}
		\caption{Visualization of heat maps. AU stands for the action units defined in facial action coding system (FACS) \cite{ekman1978facs}.}\vspace{-0.3cm}
		\label{fid:cam}
	\end{minipage}
\end{figure}

\subsection{Visualization}
\subsubsection{Reconstruction Results in Pre-training Stage: }To demonstrate MA2MI more intuitively, we visualize the reconstruction results in pre-training process. From Fig.~\ref{fig:rec}, it can be seen that DiT can effectively reconstruct $I_{t+\delta}$ based on $I_{t}$ and $C_\Delta$. Specifically, in the first line, $\hat{I}_{t+\delta}$ reconstructs the small action of the mouth. This demonstrates that $C_\Delta$ can accurately present the subtle movements between two frames, which is crucial for MER.

\subsubsection{Visualization of the Heat Maps: } We also show the heat maps through Grad-CAM \cite{selvaraju2017grad} in Fig.~\ref{fid:cam}. The results show that the region of interest of MIACNet is highly correlated with the region where action occurs between the onset and apex frames. Besides, these regions of interest correspond to the action unit annotations. For example, R1 and R2 indicate the right inner brow raiser and right outer brow raiser, which corresponds to the heat map of the second row.

\section{Conclusion}
In this work, we propose a transfer learning paradigm, named MA2MI. Under MA2MI, the network can be trained through reconstruction task and does not require any manual annotations of macro-expression data, which makes our method have a wider applicability. Besides, we devise micro-action network that can decouple facial position and facial action features through two independent encoders. These two branches are trained independently with different losses in the pre-training stage, which allows facial actions to be located to specific facial areas. MA2MI can achieve state-of-the-art performance on different MER datasets by pre-training on macro-expression datasets.

\bibliographystyle{splncs04}
\bibliography{main}
\end{document}